\documentclass[letterpaper, 10 pt, conference]{ieeeconf}  

\IEEEoverridecommandlockouts                              

\usepackage{cite}
\usepackage{amsmath,amssymb}
\usepackage{graphicx}
\usepackage{adjustbox}
\usepackage{textcomp}
\usepackage{hyperref}
\let\labelindent\relax
\usepackage{enumitem}
\usepackage[font=small,labelfont=bf]{caption}
\usepackage{xcolor}
\usepackage{booktabs}
\usepackage{multirow}
\usepackage{subfiles}
\usepackage{pifont}

\newcommand{\cmark}{\textcolor[HTML]{009933}{\ding{52}}}  
\newcommand{\xmark}{\textcolor[HTML]{CC0000}{\ding{56}}}  

\title{\LARGE \bf
RangeNav: Task-Aligned LiDAR Range-Image Pretraining \\
for Robot Navigation
}

\newif\ifanonymous
\anonymousfalse 

\DeclareRobustCommand{\anontext}[2]{%
  \ifanonymous
    \textcolor{orange}{[#2]}%
  \else
    #1%
  \fi
}

\DeclareRobustCommand{\anon}[1]{%
  \ifanonymous
  \else
    #1%
  \fi
}

\author{
\anontext{
Samuel Foo Enze$^{1}$,
Manthan Patel$^{1}$,
Fan Yang$^{1}$,
Linus Kramer$^{1}$,
Pascal Roth$^{1,2}$,
Marco Hutter$^{1}$,
Cesar Cadena$^{1}$
\thanks{$^1$ Robotic Systems Lab (RSL), ETH Zurich, Switzerland}
\thanks{$^2$ NVIDIA}
}{anonymous authors}
}

\begin{document}
\maketitle
\thispagestyle{empty}
\pagestyle{empty}

\begin{abstract}
    Autonomous mobile robots operating in dense crowds and cluttered sites require $360^\circ$ perception to react to surrounding hazards. While panoramic LiDAR offers full-surround coverage, learning representations within reinforcement learning (RL) is sample-inefficient under sparse rewards, and existing 3D pretrained point-cloud encoders are too memory-heavy for massively parallel RL simulation. We present \textbf{RangeNav}, a compact (183k parameters) LiDAR range-image encoder pretrained entirely on synthetic data. RangeNav uses anisotropic convolutions and circular padding to handle the extreme aspect ratio and azimuthal seam of range images. We pretrain the network using a navigation-aligned objective combining range reconstruction, semantic segmentation, and bird's-eye-view (BEV) prediction, then freeze it for policy learning. In simulation, adding semantic and BEV supervision increases success rates over a reconstruction-only baseline by an average of 32.03 and 23.73 percentage points (pp) in procedurally generated training and out-of-distribution realistic test environments, respectively. Zero-shot real-world deployments on a Unitree B2-W quadruped demonstrate robust navigation through dense crowds, low-light settings, and construction zones where limited-field-of-view depth policies struggle. Code and models will be open-sourced. Project website: \url{https://rangenav.github.io/}.
\end{abstract}

\section{Introduction}
\label{sec:introduction}

Autonomous navigation enables mobile robots to operate in challenging real-world domains, from negotiating dense pedestrian crowds in last-mile delivery to inspecting unstructured industrial and construction sites. A major driver of recent progress in learning-based navigation has been the use of pretrained visual representations. Rather than requiring the policy to learn perception from scratch, a frozen or lightly fine-tuned encoder supplies rich geometric and semantic features. This approach has proven effective for RGB-based navigation~\cite{cai2025navdp,inglin2026less} and depth image-based policies~\cite{yang2025spatially,patel2026defm}.

However, both RGB and depth cameras inherit limitations. RGB features degrade under poor lighting and appearance shifts, and depth cameras return noisy or missing values beyond a few meters, in sunlight, and on reflective or textureless surfaces. Crucially, both are limited to a narrow field of view (FoV). Obstacles approaching from the side, pedestrians overtaking from behind, and routes outside the current view are simply invisible until it is too late to react. LiDAR addresses these weaknesses with $360^\circ$ coverage, metric range up to tens of meters, and invariance to illumination, making it the natural sensor for the crowded, outdoor, and industrial settings that motivate this work. Yet pretrained representations for LiDAR-based navigation remain underexplored. Although point-cloud foundation models (FMs) such as Sonata~\cite{wu2025sonata} and Utonia~\cite{zhang2026utonia} learn strong geometric and semantic features, their computational cost limits their use in massively parallel reinforcement learning (RL), and their effectiveness as pretraining sources for compact navigation encoders remains an open question. A navigation-ready LiDAR encoder should therefore be \emph{lightweight} enough for large-scale RL, \emph{pretrained} so the policy does not have to learn perception from sparse rewards, and \emph{generalizable} across scenes and to the real world.

\begin{figure}[t]
    \centering
    \includegraphics[width=\columnwidth]{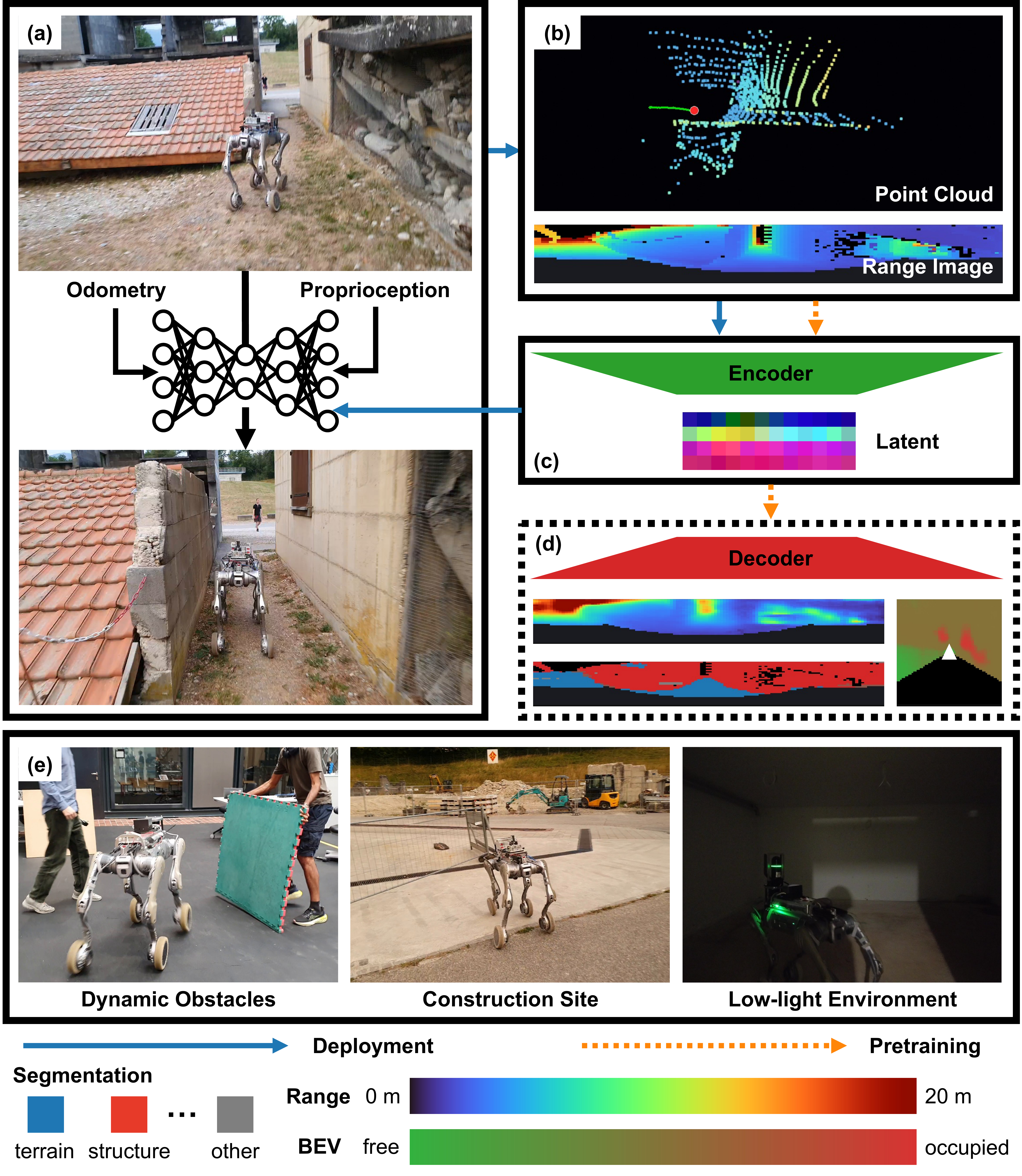}
    \caption{System overview. (a) A mobile robot observes the environment with LiDAR, odometry, and proprioception. (b) The LiDAR scan is projected into a panoramic range image. (c) Our pretrained encoder compresses the range image into a spatial latent for the navigation policy. (d) During pretraining, task-aligned decoders supervise range reconstruction, semantic segmentation, and bird's-eye-view (BEV) prediction. (e) The frozen encoder transfers to real-world navigation in dynamic environments, construction sites, and low-light environments without real-world training.}
    \label{fig:overview}
\end{figure}

To this end, we present a compact LiDAR range-image encoder pretrained specifically for navigation. Exploiting the fixed scan pattern of spinning LiDARs, we spherically project each scan onto a 2D \emph{range image}~\cite{milioto2019rangenet}. This representation is cheap to compute and batches predictably across parallel environments, but its geometry differs from a perspective image: it has many more azimuthal columns than elevation rows, and the image wraps around at the $0^\circ/360^\circ$ seam. We account for both with a fully convolutional encoder that uses anisotropic kernels and circular convolutions along the azimuth. We pretrain the encoder on range images rendered from
TartanGround~\cite{patel2025tartanground} with three navigation-aligned objectives: range reconstruction, semantic segmentation, and BEV prediction, which preserve object semantics, obstacle boundaries, and top-down traversability information. The frozen latent representation then feeds an RL navigation policy using spatially-enhanced recurrent units (SRUs)~\cite{yang2025spatially} (Fig.~\ref{fig:overview}).

In simulation, our pretraining recipe improves navigation success rate (SR) over a reconstruction-only baseline by an average of 32.03 and 23.73 percentage points (pp) in procedurally generated training and out-of-distribution test environments, respectively. Compared with the depth-based policy of~\cite{yang2025spatially}, our LiDAR policy achieves comparable average SR, shorter navigation times, and higher success weighted by path length (SPL), and surpasses it in crowded and densely cluttered environments, highlighting the benefits of a wider FoV. Deployed zero-shot on a Unitree B2-W robot, the policy navigates obstacle-rich environments, unlit spaces, dense crowds, and cluttered construction sites, settings in which limited-FoV depth policies struggle.

In summary, our contributions are:
\begin{enumerate}[leftmargin=*,nosep]
    \item A navigation-aligned pretraining recipe combining range reconstruction, semantic segmentation, and BEV prediction that significantly improves downstream navigation.
    \item A lightweight (183k parameters) LiDAR range-image encoder with anisotropic and circular convolutions, suitable for large-scale RL.
    \item Extensive simulation and zero-shot real-world evaluation demonstrating the advantages of LiDAR-based navigation policies over limited-FoV depth-based policies.
\end{enumerate}

\section{Related Work}
\label{sec:related}

\paragraph{Pretrained perception for navigation}

Recent methods decouple perception learning from downstream control, reducing the perceptual burden on policy learning caused by sparse rewards~\cite{majumdar2023we, stooke2021atc, yang2025spatially}. Several works build upon RGB vision foundation models~\cite{oquab2023dinov2, ranzinger2024radio, ravi2025sam}, keeping encoders frozen or lightly fine-tuned for point-goal navigation~\cite{cai2025navdp, inglin2026less}, image-goal navigation~\cite{majumdar2023we,zhou2024dino}, and visual affordance prediction~\cite{frey2023fast, schmittle2025lrn, shah2026wildos}. Concurrently, representations learned directly from depth images~\cite{yang2023iplanner, yang2025spatially, patel2026defm} have demonstrated robust zero-shot sim-to-real transfer for navigation. While these camera-based representations highlight the benefits of pretrained perception, they are typically designed for perspective projections with dense, isotropic pixel grids. Adapting them to panoramic LiDAR range images requires accounting for the large horizontal-to-vertical aspect ratio, structured beam sparsity, and azimuthal wraparound.

\paragraph{LiDAR representations}

Most general-purpose LiDAR encoders operate directly on irregular 3D geometry. Seminal architectures like PointNet++~\cite{qi2017pointnetplusplus} and MinkowskiEngine~\cite{choy20194d} consume raw point sets or sparse spatial tensors, while recent FMs such as Sonata~\cite{wu2025sonata} and Utonia~\cite{zhang2026utonia} pretrain Point Transformers~\cite{wu2024point} across large-scale 3D datasets. Lightweight variants like LitePT~\cite{yue2026litept} optimize hierarchical point processing to reduce overhead. However, point-based and sparse-tensor operations incur prohibitive memory and runtime costs across thousands of parallel RL environments (Sec.~\ref{sec:experiments}). Projection-based approaches circumvent these costs by structuring scans into standard arrays. Notably, RangeNet++~\cite{milioto2019rangenet} showed that spherical projection allows efficient 2D convolutions, and RangeFormer~\cite{kong2023rangeformer} further scaled range representations using vision transformers. However, these systems are optimized and benchmarked for semantic perception, and pretrained range encoders remain largely unexplored for RL navigation.

\paragraph{Navigation-aligned representation learning}

Effective pretraining preserves task-relevant information. Prior work guides visual navigation encoders with auxiliary supervision, including action and temporal-distance predictions~\cite{ye2021auxiliary}, dynamics and progress targets~\cite{desai2021auxiliary}, depth prediction and loop-closure classification~\cite{mirowski2017navigation}, and auxiliary occupancy prediction for stereo cameras~\cite{li2024stereonavnet}. While these studies show the benefit of task-aligned supervision for camera inputs, combining complementary geometric and semantic objectives for LiDAR navigation has received little attention.

\section{Method}
\label{sec:method}

Our approach consists of three stages. We first project synthetic depth observations into LiDAR range images to generate a large pretraining dataset. We then pretrain a deterministic autoencoder with reconstruction, semantic segmentation, and BEV prediction objectives. Finally, we train a recurrent actor--critic navigation policy utilizing the spatial latents obtained from the frozen encoder.

\begin{figure}[t]
    \centering
    \includegraphics[width=\columnwidth]{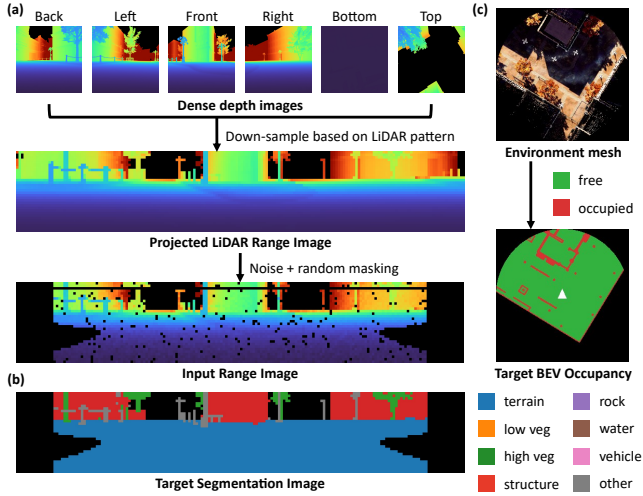}
    \caption{Data processing for encoder pretraining. (a) Six dense depth views are fused in 3D and projected onto the target spherical LiDAR grid, then corrupted with noise and random masking to form the encoder input. (b) A similar approach generates the semantic target from segmented images. (c) The privileged BEV occupancy and ground-height target is rendered from the scene mesh.}
    \label{fig:data_processing}
\end{figure}

\subsection{Range-Image Generation}

We use TartanGround~\cite{patel2025tartanground} to generate synthetic LiDAR range images. For each pose, six depth images with known intrinsics and extrinsics are back-projected and transformed into a common sensor frame. The aggregated point cloud is mapped into the azimuth and elevation bins of a target LiDAR's spherical range image (Fig.~\ref{fig:data_processing}(a)), with bin collisions resolved by minimum pooling. Projecting 3D point clouds into 2D range images yields fixed-shape tensors for efficient batching across thousands of parallel RL environments at lower computational cost than 3D point-cloud encoders (Table~\ref{tab:scalability_status}).

\begin{figure}[t]
    \centering
    \includegraphics[width=\columnwidth]{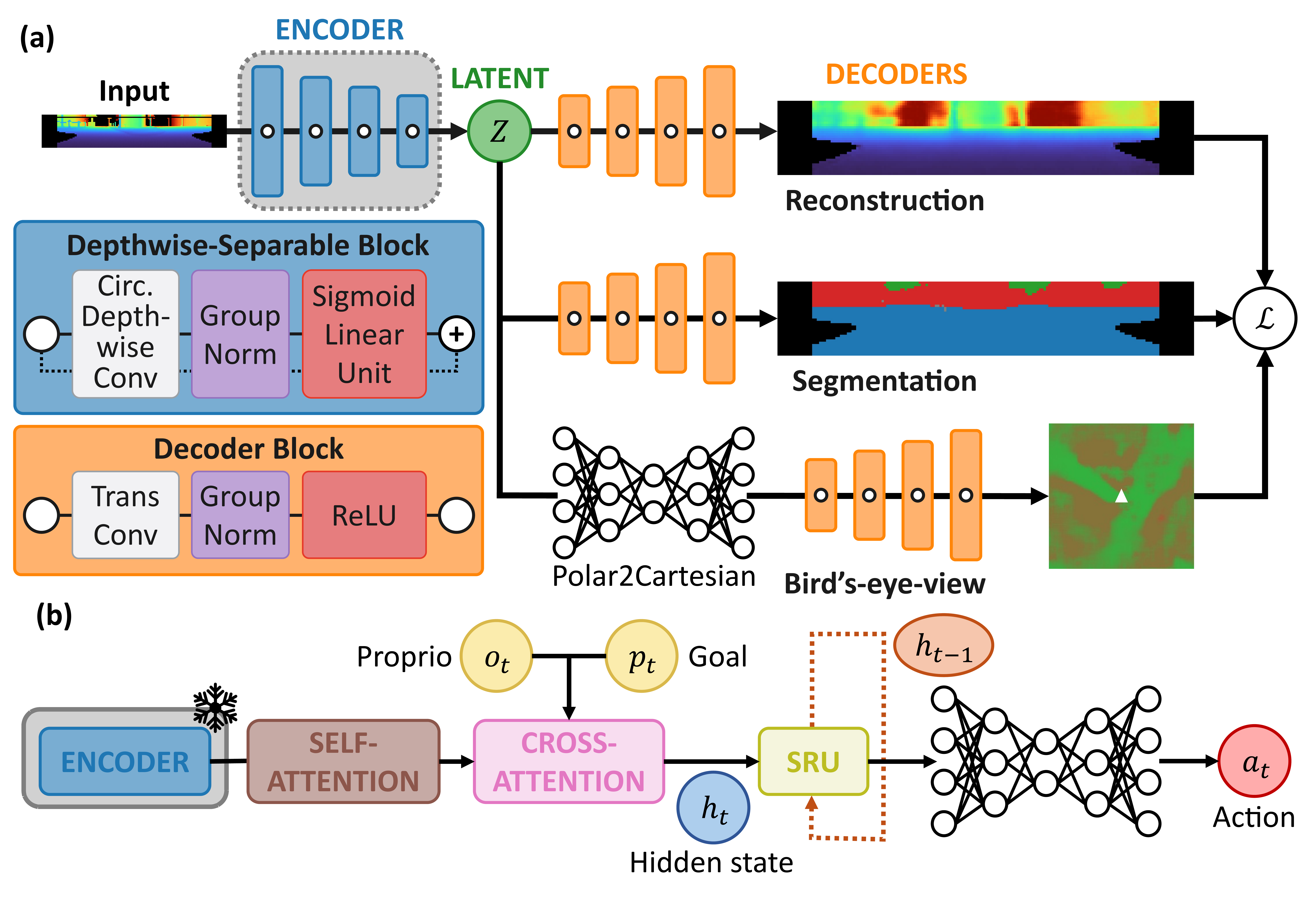}
    \caption{Pretraining and policy architecture. (a) The encoder compresses the range image into a spatial latent that three decoder heads read in parallel: range reconstruction, per-return semantic segmentation, and privileged BEV occupancy and ground height. A fully connected bottleneck precedes the BEV deconvolutions, mapping the polar latent onto the Cartesian grid. Insets detail the depthwise-separable encoder block and the transposed-convolution decoder block. All heads are discarded after pretraining. (b) The frozen encoder feeds attention over noisy proprioception and the relative goal into the recurrent SRU policy that predicts the action.}
    \label{fig:training_architecture}
\end{figure}

\subsection{Range Preprocessing and Augmentation}
\label{sec:range_preprocessing}

Let $r$ denote a clean metric range image. During encoder pretraining, we apply random circular horizontal rolls and corrupt valid ranges with scan-ring bias, distance- and edge-dependent Gaussian noise, random dropout, and contiguous masking. The corresponding clean, rolled scan remains the reconstruction target. These augmentations simulate arbitrary sensor yaw, measurement noise, missing returns, and occlusion (Fig.~\ref{fig:data_processing}(a)). For a clean measurement $r_{ij}$, the pre-dropout corrupted measurement is
\begin{equation}
    \tilde r_{ij}=\operatorname{clip}(r_{ij}+b_i+\eta_{ij},d_{\min},d_{\max}),
\end{equation}
where $b_i\sim\mathcal{N}(0,\sigma_{\mathrm{ring}}^2)$ is shared by scan ring $i$, and the variance of $\eta_{ij}$ grows with range and local edge magnitude. The dropout probability likewise increases with range.

Before encoding, invalid returns in $r$ and $\tilde r$ are filled with the maximum working range and all ranges are clipped to the LiDAR's working interval. Finally, we log-transform the range images as $x_{ij}=\log(1+r_{ij})$ and $\tilde{x}_{ij}=\log(1+\tilde r_{ij})$. The logarithm emphasizes nearby geometry while compressing distant returns. We use $\tilde{x}$ as the encoder input and $x$ as its corresponding clean reconstruction target.

\subsection{Task-Aligned Pretraining}

We pretrain the encoder on range images with geometric and semantic targets, allowing the pretraining dataset to scale through scene rendering without requiring action sequences or goal-conditioned rollouts. Within the encoder (Fig.~\ref{fig:training_architecture}(a)), circular convolutions preserve continuity across the $360^\circ$ azimuthal seam~\cite{yang2024tulip}, while depthwise-separable residual blocks reduce computational cost and anisotropic convolutions provide wider horizontal receptive fields suited to the panoramic range-image grid~\cite{wu2018squeezeseg,yang2024tulip}. The encoder produces the latent $z_L$ and applies layer normalization~\cite{ba2016layer} independently at each spatial location.

Three decoder heads condition this bottleneck. The range decoder predicts the clean log-transformed range image $\hat{x}$, the semantic head predicts a semantic label for each valid return, and the BEV head predicts top-down occupancy and ground height. The combined objective is
\begin{equation}
    \begin{aligned}
        \mathcal{L}_{\mathrm{pre}} ={} &
        \lambda_{\mathrm{recon}}\mathcal{L}_{\mathrm{recon}}
        +\lambda_{\mathrm{grad}}\mathcal{L}_{\mathrm{grad}}                                \\
                                       & +\lambda_{\mathrm{seg}}\mathcal{L}_{\mathrm{seg}}
        +\lambda_{\mathrm{BEV}}\mathcal{L}_{\mathrm{BEV}} .
    \end{aligned}
    \label{eq:pretrain}
\end{equation}
The $\lambda$ coefficients weight their corresponding objectives. The range decoder is supervised jointly by $\mathcal{L}_{\mathrm{recon}}$, an edge-weighted masked Huber loss on its prediction $\hat{x}$, and $\mathcal{L}_{\mathrm{grad}}$, which matches the spatial gradients of $\hat{x}$ and the clean target $x$. More precisely, let $e_{ij}$ denote the target range discontinuity around pixel $(i,j)$ and $\bar e$ its clamped mean. With valid mask $m_{ij}$ and edge weight $w_{ij}=1+\alpha_e e_{ij}/\bar e$, the reconstruction term is
\begin{equation}
    \mathcal{L}_{\mathrm{recon}}=
    \frac{\sum_{ij}m_{ij}w_{ij}\,
    \mathrm{Huber}(\hat x_{ij},x_{ij})}
    {\sum_{ij}m_{ij}w_{ij}}.
\end{equation}
The gradient term applies the same robust penalty to adjacent horizontal and vertical differences in $\hat{x}$ and $x$ wherever both target pixels are valid. Let $\mathcal{D}=\{(0,1),(1,0)\}$ denote the horizontal and vertical offsets and $\Delta_d x_{ij}=x_{i+d_i,j+d_j}-x_{ij}$. Then
\begin{equation}
    \mathcal{L}_{\mathrm{grad}}=
    \frac{\displaystyle\sum_{d\in\mathcal{D}}\sum_{ij}
    m_{ij}m_{i+d_i,j+d_j}\,
    \mathrm{Huber}\!\left(\Delta_d\hat{x}_{ij},\Delta_d x_{ij}\right)}
    {\displaystyle\sum_{d\in\mathcal{D}}\sum_{ij}
    m_{ij}m_{i+d_i,j+d_j}}.
\end{equation}
Semantic labels are projected with the same spherical bins and the raw TartanGround labels are collapsed into eight classes: terrain, low vegetation, high vegetation, structure, rock, water, vehicle, and other (Fig.~\ref{fig:data_processing}(b)). On valid, non-self-occluded returns,
\begin{equation}
    \mathcal{L}_{\mathrm{seg}}=
    \mathcal{L}_{\mathrm{CE}}+\mathcal{L}_{\mathrm{Lov\acute{a}sz}}.
\end{equation}
Here, $\mathcal{L}_{\mathrm{CE}}$ is the mean pixel-wise cross-entropy loss, and $\mathcal{L}_{\mathrm{Lov\acute{a}sz}}$ is the Lov\'asz--Softmax loss, a differentiable surrogate for the mean intersection-over-union (mIoU) segmentation metric~\cite{berman2018lovasz}.
The privileged BEV target contains free/blocked occupancy and ground height on a Cartesian grid, supervised from scene-level semantic point clouds in visible and occluded cells (Fig.~\ref{fig:data_processing}(c)). For known cells,
\begin{equation}
    \mathcal{L}_{\mathrm{BEV}}=\mathcal{L}_{\mathrm{CE}}^{\mathrm{occ}}+
    \mathcal{L}_{\mathrm{Lov\acute{a}sz}}^{\mathrm{occ}}+
    \mathcal{L}_{1}^{\mathrm{height}}.
\end{equation}
The term $\mathcal{L}_{1}^{\mathrm{height}}$ is the mean absolute error between the predicted and target ground heights over known cells.

All three decoder heads are used only during pretraining; only the encoder is retained for the RL policy.

\subsection{Recurrent Navigation Policy}

The pretrained LiDAR encoder is frozen during policy learning (Fig.~\ref{fig:training_architecture}(b)). Its spatial latent at time $t$, $z_{L,t}$, is combined with noisy proprioception and the relative goal before entering the recurrent memory:
\begin{equation}
    h_t = \mathrm{SRU}(z_{L,t},o_t^{\mathrm{prop}},g_t,h_{t-1}).
\end{equation}
Here, $o_t^{\mathrm{prop}}$ is the proprioceptive observation, $g_t$ is the relative goal, and $h_t$ and $h_{t-1}$ are the current and previous hidden states, respectively.
The actor predicts the control action from $h_t$, while an asymmetric critic can use privileged geometric observations during training.

We train the actor--critic with Mutual Distillation Policy Optimization (MDPO)~\cite{xie2025representation} in massively parallel Isaac Lab environments. Randomized terrain, dynamics, sensing, actuation, delays, and disturbances expose the policy to transfer-relevant variation. Episodes end on goal reach, timeout, unsafe contact, excessive roll/pitch, or a fall. Following SRU~\cite{yang2025spatially}, the reward combines sparse task rewards for reaching the goal with dense shaping rewards for goal progress, smooth motion, and path efficiency.

\section{Experiments}
\label{sec:experiments}

We first outline the implementation details (Sec.~\ref{sec:implementation_details}) and evaluation protocol (Sec.~\ref{sec:eval_prot}). Next, our experiments evaluate three central questions: (i) how do our pretraining objectives affect navigation SR (Sec.~\ref{sec:pretraining_objectives}), (ii) how does panoramic LiDAR compare with front-facing depth (Sec.~\ref{sec:lidar_vs_depth}), and (iii) how does our range-image encoder compare with general-purpose point-cloud encoders (Sec.~\ref{sec:point_cloud_comparison}). We then ablate the design choices of our encoder (Sec.~\ref{sec:encoder_ablations}), and finally perform several real-world deployments (Sec.~\ref{sec:real_world_deployments}).

\subsection{Implementation Details}
\label{sec:implementation_details}
For encoder pretraining, six TartanGround depth views are fused and sampled to simulate a 32-beam RoboSense Helios LiDAR with a $70^\circ$ vertical field of view (VFoV) and a $360^\circ$ horizontal field of view (HFoV).
We use a $0.4^\circ$ horizontal resolution and a range of $[0.2,20]$\,m. After adding noise to the resulting $32\times900$ range image, we apply minimum pooling along the azimuth to reduce it to $32\times180$, preserving the closest returns while eliminating horizontal redundancy. For the noise model in Sec.~\ref{sec:range_preprocessing}, we use $\sigma_0=0.01$, range factor $0.002$, edge factor $0.03$, base dropout $0.005$, range-dependent dropout factor $0.02$, and $\sigma_{\mathrm{ring}}=0.01$.

The encoder maps an input in $\mathbb{R}^{1\times32\times180}$ to a spatial latent in $\mathbb{R}^{64\times4\times12}$. The pretraining dataset contains around 144,000 frames. We jointly optimize the encoder and all enabled decoder heads for 5,000 epochs using the Adam optimizer with a batch size of 1,024 and a learning rate of $6\times10^{-4}$. The loss weights are set to $\lambda_{\mathrm{recon}}=1$, $\lambda_{\mathrm{grad}}=0.3$, $\lambda_{\mathrm{seg}}=2$, and $\lambda_{\mathrm{BEV}}=2$ when the corresponding heads are enabled, with edge-weight coefficient $\alpha_e=0.5$. The BEV target covers $40\times40$\,m with a $256\times256$ grid at 0.15625\,m per cell.

For the RL training stage, we train the actor--critic with MDPO in 1,920 parallel environments. The policy has a one-layer 512-unit recurrent state followed by $[512,256,128]$ actor and critic multilayer perceptrons (MLPs). The policies in Table~\ref{tab:objective_results} are trained for 15,000 iterations, whereas Tables~\ref{tab:encoder_ablation} and~\ref{tab:pretraining_objectives} use a shorter budget of 5,000 training iterations for ablations. Maximum episode duration ranges from 60 to 120\,s across environments.

To increase the robustness of the encoder and RL policy, we randomly mask parts of the range image to simulate occlusions by different robot-body and payload patterns. We also add stick noise to simulate returns from thin poles, bars, and ring-like structures near the sensor. These patterns model persistent structured artifacts that are not captured by independent Gaussian noise or dropout.

\subsection{Evaluation Protocol}
\label{sec:eval_prot}
\begin{figure*}[t]
    \centering
    \includegraphics[width=\textwidth]{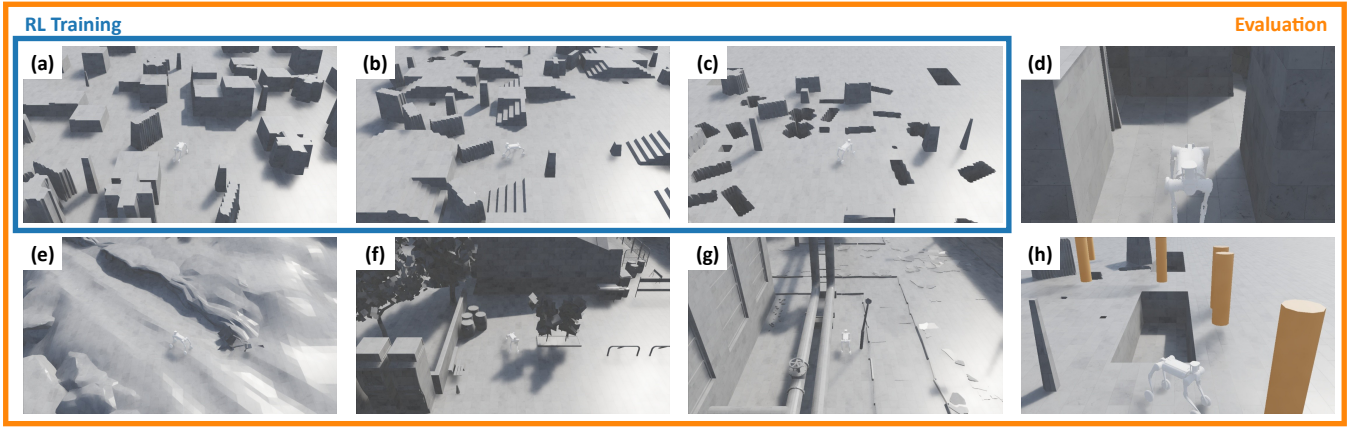}
    \caption{Simulation environments. (a) Maze, (b) stairs, (c) pits, (d) side wall, (e) Gascola, (f) ModernCityDowntown (City), (g) AbandonedCable (Cable), and (h) dynamic obstacles, where the orange cylinders are moving pedestrians. For the results in Table \ref{tab:objective_results}, the environments in the blue box (a)--(c) are examples of the environments used for RL training, whereas all environments are used for evaluation.}
    \label{fig:sim_envs}
\end{figure*}

Fig.~\ref{fig:sim_envs} shows representative simulation environments. Success denotes reaching the goal within 0.5\,m before termination. Every evaluation uses 1,024 fixed start--goal scenarios per environment. The base mixture comprises procedurally generated mazes, pillar fields, stairs, and pits, exposing the policy to corridor selection, isolated obstacles, elevation changes, and negative obstacles. Only these environments are used for RL training. The held-out TartanGround scenes~\cite{patel2025tartanground} comprise AbandonedCable, Gascola, ModernCityDowntown, and Market, which provide more diverse layouts and realistic object arrangements. We additionally use two targeted held-out settings. The side-wall evaluation places the robot approximately 0.10\,m from walls to test navigation from cluttered starts with limited local clearance. The dynamic-obstacle evaluation introduces 13 pedestrians moving along random paths between spawn and goal locations.
For the comparison with a front-facing depth camera policy (Sec.~\ref{sec:lidar_vs_depth}), we retrain the policy from ~\cite{yang2025spatially} under comparable RL settings.

\subsection{Task-Aligned Pretraining Objectives}
\label{sec:pretraining_objectives}

We compare three cumulative LiDAR pretraining objectives: reconstruction alone, adding semantic segmentation, and adding BEV prediction. Checks in Table~\ref{tab:objective_results} mark each encoder's pretraining heads.

\begin{table*}[t]
    \caption{Navigation SR (\%) ($\uparrow$) comparing depth and LiDAR policies trained using different pretraining objectives.}
    \label{tab:objective_results}
    \centering
    \scriptsize
    \setlength{\tabcolsep}{3.5pt}
    \begin{tabular}{lccccccc|ccccc|ccc}
        \toprule
              & \multicolumn{3}{c}{Pretraining heads} & \multicolumn{5}{c}{Base mixture} & \multicolumn{5}{c}{TartanGround} & \multicolumn{1}{c}{Side-wall} & \multicolumn{1}{c}{Dynamic}                                                                                                                                                                                             \\
        \cmidrule(lr){2-4} \cmidrule(lr){5-9} \cmidrule(lr){10-14} \cmidrule(lr){15-15} \cmidrule(lr){16-16}
        Input & Recon                                 & Seg                              & BEV                              & Maze                          & Pillars                     & Stairs & Pits           & \multicolumn{1}{|c}{Average} & Cable & Gascola        & City           & Market         & \multicolumn{1}{|c}{Average} & \multicolumn{1}{c}{-} & -              \\
        \midrule
        Depth & \multicolumn{3}{c}{---}               & 95.61                            & \textbf{99.02}                   & \textbf{95.70}                & 91.11                       & 95.36  & \textbf{88.57} & 47.27                        & 83.11 & 72.85          & 72.95          & 79.98          & 77.64                                                                 \\
        \addlinespace
        LiDAR & \cmark                                & \xmark                           & \xmark                           & 78.81                         & 80.08                       & 68.07  & 26.76          & 63.43                        & 51.46 & 29.98          & 46.39          & 69.92          & 49.44                        & 61.52                 & 63.18          \\
        LiDAR & \cmark                                & \cmark                           & \xmark                           & 95.70                         & 96.09                       & 80.08  & 82.13          & 88.50                        & 77.83 & 43.07          & 80.76          & 68.07          & 67.43                        & 85.94                 & 78.03          \\
        LiDAR & \cmark                                & \cmark                           & \cmark                           & \textbf{96.88}                & 98.54                       & 94.34  & \textbf{92.09} & \textbf{95.46}               & 81.45 & \textbf{51.56} & \textbf{86.13} & \textbf{73.54} & \textbf{73.17}               & \textbf{88.87}        & \textbf{79.49} \\
        \bottomrule
    \end{tabular}
\end{table*}

The encoder supervised by segmentation and BEV objectives is the strongest LiDAR variant in every column. Notably, it improves substantially over reconstruction on Pits (26.76\% to 92.09\%) and AbandonedCable (51.46\% to 81.45\%), environments with many obstacles below the sensor height, where the LiDAR beams are sparser. A reconstruction objective can be dominated by the many easier returns without ensuring that the latent preserves these relatively small but safety-critical structures. Semantic supervision encourages the encoder to distinguish geometrically similar returns with different navigation implications. BEV supervision then provides a further gain by organizing those features around free space and terrain layout. This improvement in navigation performance persists outside the RL training mixture as well, indicating that the benefit transfers across layouts. Together, the results suggest that semantic and BEV supervision addresses the challenges of sparse LiDAR data and produces a more navigation-aligned LiDAR latent than reconstruction alone.

\subsection{Panoramic LiDAR versus Front-Facing Depth}
\label{sec:lidar_vs_depth}

Next, we compare our LiDAR policy with the dense front-facing depth policy to evaluate whether our encoder provides effective navigation features from sparse LiDAR observations and whether panoramic coverage offers advantages over denser observations restricted to the robot's forward view.

Across the base mixture and TartanGround environments, our LiDAR policy achieves higher SR than the depth policy in five of the eight environments (Table~\ref{tab:objective_results}). Depth achieves higher SR on stairs, pillars, and Cable. Its advantage on stairs and Cable may reflect denser observations of step geometry and thin or low-lying obstacles that sparse LiDAR beams can miss. LiDAR performs better in the remaining environments, including pits with negative obstacles and mazes with corridor layouts, with the largest gains in Gascola and City, which feature uneven outdoor terrain and urban street layouts. Overall, our LiDAR policy achieves slightly higher average SR on both evaluation sets despite its sparser observations, suggesting that our encoder provides features useful for navigation.

Front-facing depth also creates a partial-observability problem when safe navigation depends on geometry outside the robot's current viewing direction. This limitation is particularly evident in two evaluation settings (Table~\ref{tab:objective_results}). In the side-wall setting, the robot starts closer to walls. As it turns, the robot may collide with surrounding obstacles outside of the frontal depth view, leading to the depth policy failing more frequently. Panoramic LiDAR retains this lateral geometry and enables the policy to account for the available clearance before moving sideways or turning, improving SR from 79.98\% to 88.87\%. In the dynamic-obstacle setting, pedestrians can approach from either side and enter the robot's path before appearing in the frontal view. By observing these agents earlier, the LiDAR policy can react and avoid them more reliably, increasing SR from 77.64\% to 79.49\% (Table~\ref{tab:objective_results}).

\begin{figure}[t]
    \centering
    \includegraphics[width=\columnwidth]{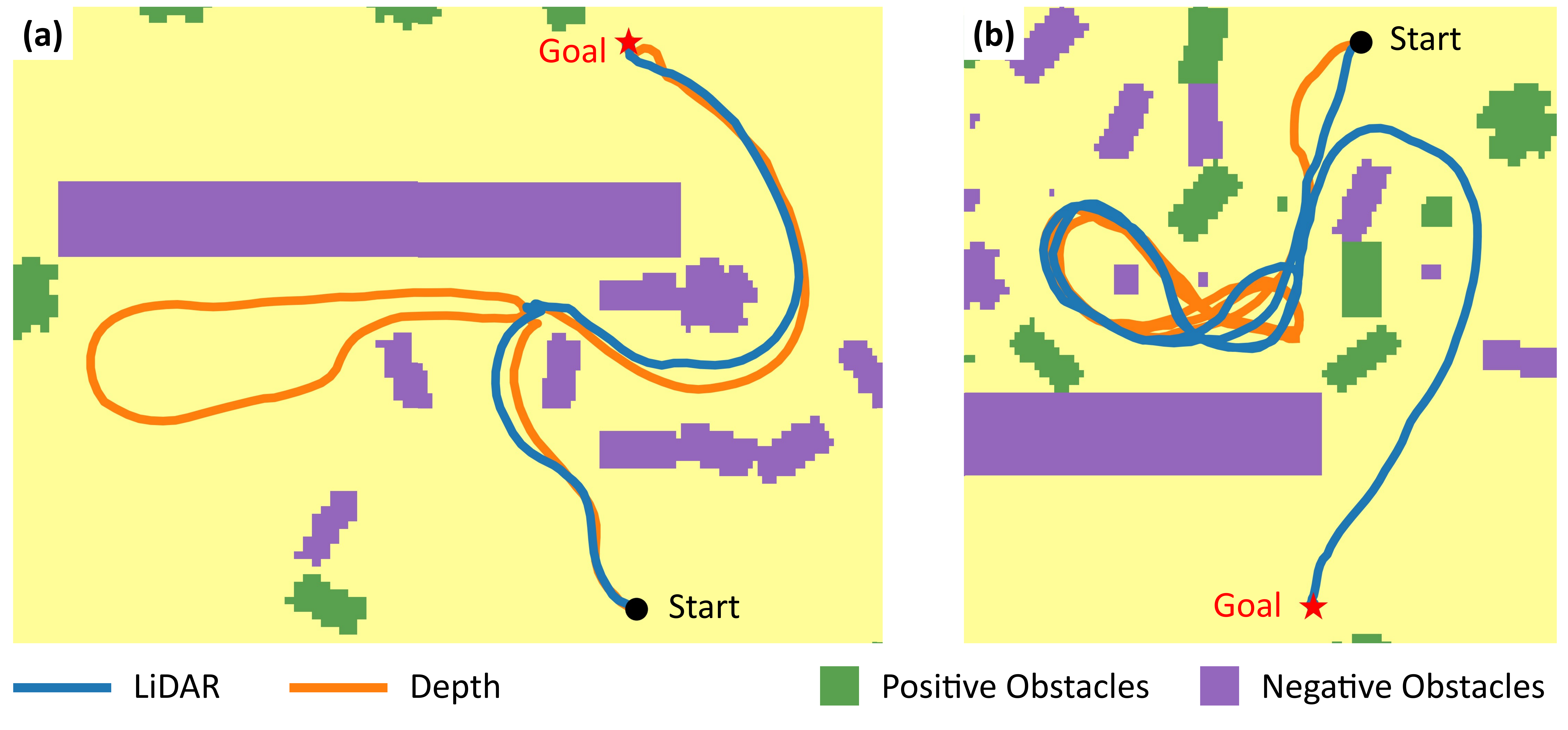}
    \caption{Representative LiDAR (blue) and depth (orange) policy trajectories. (a) The depth policy detours into a dead end before recovering, whereas the LiDAR policy recognizes it earlier and takes a direct route. (b) The depth policy repeatedly revisits the same local clutter, whereas the LiDAR policy escapes and reaches the goal.}
    \label{fig:trajectory_comparison}
\end{figure}

Beyond success rate, panoramic context improves route efficiency when multiple passages are locally plausible. This effect is particularly evident in the maze setting, where LiDAR can reveal lateral openings and surrounding corridor structure without requiring the robot to reorient or enter an unproductive passage. Across the base-mixture environments, the LiDAR policy consequently achieves an SPL score 0.58\,pp higher than the depth policy and reaches the goal 1.38\,s faster on average (Table~\ref{tab:lidar_depth_path_efficiency}). The representative trajectories in Fig.~\ref{fig:trajectory_comparison} illustrate this behavior: the LiDAR policy recognizes a dead end earlier and avoids repeatedly revisiting clutter, producing a more direct route.

\begin{table}[t]
    \caption{LiDAR and depth navigation efficiency.}
    \label{tab:lidar_depth_path_efficiency}
    \centering
    \scriptsize
    \setlength{\tabcolsep}{3pt}
    \begin{adjustbox}{max width=\columnwidth}
        \begin{tabular}{lrr@{\hspace{7pt}}rr}
            \toprule
                    & \multicolumn{2}{c}{SPL (\%, $\uparrow$)} & \multicolumn{2}{c}{Time to goal (s, $\downarrow$)}                          \\
            \cmidrule(lr){2-3} \cmidrule(lr){4-5}
            Split   & LiDAR                                    & Depth                                              & LiDAR          & Depth \\
            \midrule
            Maze    & \textbf{90.15}                           & 89.37                                              & \textbf{15.87} & 17.94 \\
            Pillars & \textbf{93.71}                           & 92.50                                              & \textbf{11.89} & 12.90 \\
            Stairs  & \textbf{75.38}                           & 75.33                                              & \textbf{16.41} & 17.82 \\
            Pits    & \textbf{81.19}                           & 80.91                                              & \textbf{16.60} & 17.60 \\
            \midrule
            All     & \textbf{85.11}                           & 84.53                                              & \textbf{15.19} & 16.57 \\
            \bottomrule
        \end{tabular}
    \end{adjustbox}
\end{table}

\subsection{Comparison with General-Purpose Point-Cloud Encoders}
\label{sec:point_cloud_comparison}

Next, we compare our pretraining objectives with recent cross-domain and general-purpose point-cloud encoders Sonata~\cite{wu2025sonata} and Utonia~\cite{zhang2026utonia}.

Firstly, we measure the computational efficiency of the encoders by comparing the maximum number of parallel RL environments that fit in GPU memory and the resulting training throughput on a single NVIDIA RTX 4090 (Table~\ref{tab:scalability_status}). The point-cloud encoders are large, accommodate relatively few parallel environments, and reach only 19--75 steps/s during RL training, making them impractical for large-scale training. In contrast, our range-image encoder achieves 1,469 steps/s with 1,920 parallel environments, supporting efficient, massively parallel RL training.

\begin{table}[t]
    \caption{Encoder scalability for parallel RL.}
    \label{tab:scalability_status}
    \centering
    \scriptsize
    \setlength{\tabcolsep}{4pt}
    \begin{adjustbox}{max width=\columnwidth}
        \begin{tabular}{llrrr}
            \toprule
            Encoder                       & Representation         & Params        & Max RL envs ($\uparrow$) & Throughput (steps/s, $\uparrow$) \\
            \midrule
            Utonia~\cite{zhang2026utonia} & Point transformer v3   & 137M          & 48                       & 19                               \\
            Sonata~\cite{wu2025sonata}    & Point transformer v3   & 108M          & 64                       & 30                               \\
            LitePT-S~\cite{yue2026litept} & Hybrid point hierarchy & 12.7M         & 96                       & 75                               \\
            Ours                          & Fixed range image      & \textbf{183k} & \textbf{1,920}           & \textbf{1,469}                   \\
            \bottomrule
        \end{tabular}
    \end{adjustbox}
\end{table}

Nevertheless, in order to still compare with these point-cloud encoders, we distill their features into our range-image encoder architecture on the same TartanGround pretraining data. The distillation uses cosine similarity between the student and teacher features. For each teacher, we report its best-performing configuration: XYZ coordinates as input for Sonata, and XYZ coordinates, RGB color, and surface normals for Utonia. We then compare the downstream RL SR against our policy, which uses the task-aligned encoder. Table~\ref{tab:pretraining_objectives} summarizes the results.

Evaluating on the base-mixture episodes, our pretraining objective exceeds the best Sonata- and Utonia-distilled variants by 7.32\,pp and 11.40\,pp, respectively (Table~\ref{tab:pretraining_objectives}). Thus, our pretraining objective produces stronger downstream features than distillation from these point-cloud encoders. This suggests that the transferable 3D features learned by general-purpose encoders like Sonata and Utonia are not necessarily the best suited for navigation.

\begin{table}[t]
    \caption{Navigation SR (\%) ($\uparrow$) with the same encoder architecture pretrained using our pretraining objectives or features distilled from point-cloud encoders.}
    \label{tab:pretraining_objectives}
    \centering
    \scriptsize
    \setlength{\tabcolsep}{5pt}
    \begin{tabular}{llc}
        \toprule
        Pretraining source            & SR (\%) ($\uparrow$) \\
        \midrule
        Sonata~\cite{wu2025sonata}    & 85.06                \\
        Utonia~\cite{zhang2026utonia} & 80.98                \\
        Ours                          & \textbf{92.38}       \\
        \bottomrule
    \end{tabular}
\end{table}

\subsection{Encoder Architecture and Ablations}
\label{sec:encoder_ablations}

We next examine how our encoder's channel width, stem kernel size, and architecture affect navigation performance. All variants retain the same range-image representation and are evaluated on the base mixture (Table~\ref{tab:encoder_ablation}).

Firstly, to test model size, we uniformly scale our encoder's six channel widths ($32/48/64/96/128/160$) by a common factor. This scaling does not improve navigation SR.

Secondly, for kernel anisotropy, the $3\times9$ stem performs best, highlighting the importance of appropriately anisotropic kernels for LiDAR range images, whose azimuthal (width) dimension has many more pixels than their elevation (height) dimension.

\begin{table}[t]
    \caption{Range-image encoder ablations and architectural comparisons.}
    \label{tab:encoder_ablation}
    \centering
    \scriptsize
    \setlength{\tabcolsep}{2.5pt}
    \def\archrefwidth{\ensuremath{\times 1}}
    \def\archrefstem{\ensuremath{3 \times 9}}
    \begin{adjustbox}{max width=\columnwidth}
        \begin{tabular}{lccrr}
            \toprule
            Ablation                           & Channel width                                                    & Stem kernel   & Params & SR (\%) ($\uparrow$) \\
            \midrule
            Ours                               & \archrefwidth                                                    & \archrefstem  & 183k   & \textbf{92.38}       \\
            \midrule
            \multirow{3}{*}{Model size}        & $\times 1.5$                                                     & \archrefstem  & 301k   & 87.40                \\
                                               & $\times 2$                                                       & \archrefstem  & 462k   & 91.53                \\
                                               & $\times 3$                                                       & \archrefstem  & 916k   & 91.24                \\
            \midrule
            \multirow{5}{*}{Kernel anisotropy} & \archrefwidth                                                    & $3 \times 3$  & 182k   & 85.96                \\
                                               & \archrefwidth                                                    & $3 \times 5$  & 183k   & 84.67                \\
                                               & \archrefwidth                                                    & $3 \times 13$ & 183k   & 84.77                \\
                                               & \archrefwidth                                                    & $5 \times 5$  & 184k   & 80.00                \\
                                               & \archrefwidth                                                    & $5 \times 9$  & 184k   & 87.48                \\
            \midrule
            Architecture                       & \multicolumn{2}{c}{\textit{RegNet-FPN}~\cite{yang2025spatially}} & 5.8M          & 70.46  \\
            \bottomrule
        \end{tabular}
    \end{adjustbox}
    \vspace{-2mm}
\end{table}

Finally, we measure the effect of model type by comparing our encoder with the RegNet feature pyramid network (FPN) architecture~\cite{radosavovic2020regnet,yang2025spatially}. The best-performing RegNet-FPN variant obtains a lower reconstruction loss of 0.02809 than our encoder's 0.04685, yet our substantially smaller encoder obtains a much higher SR. Pixel-level range reconstruction is therefore not a reliable proxy for navigation performance.

\subsection{Real-World Deployments}
\label{sec:real_world_deployments}

\begin{figure*}[t]
    \centering
    \includegraphics[width=\textwidth]{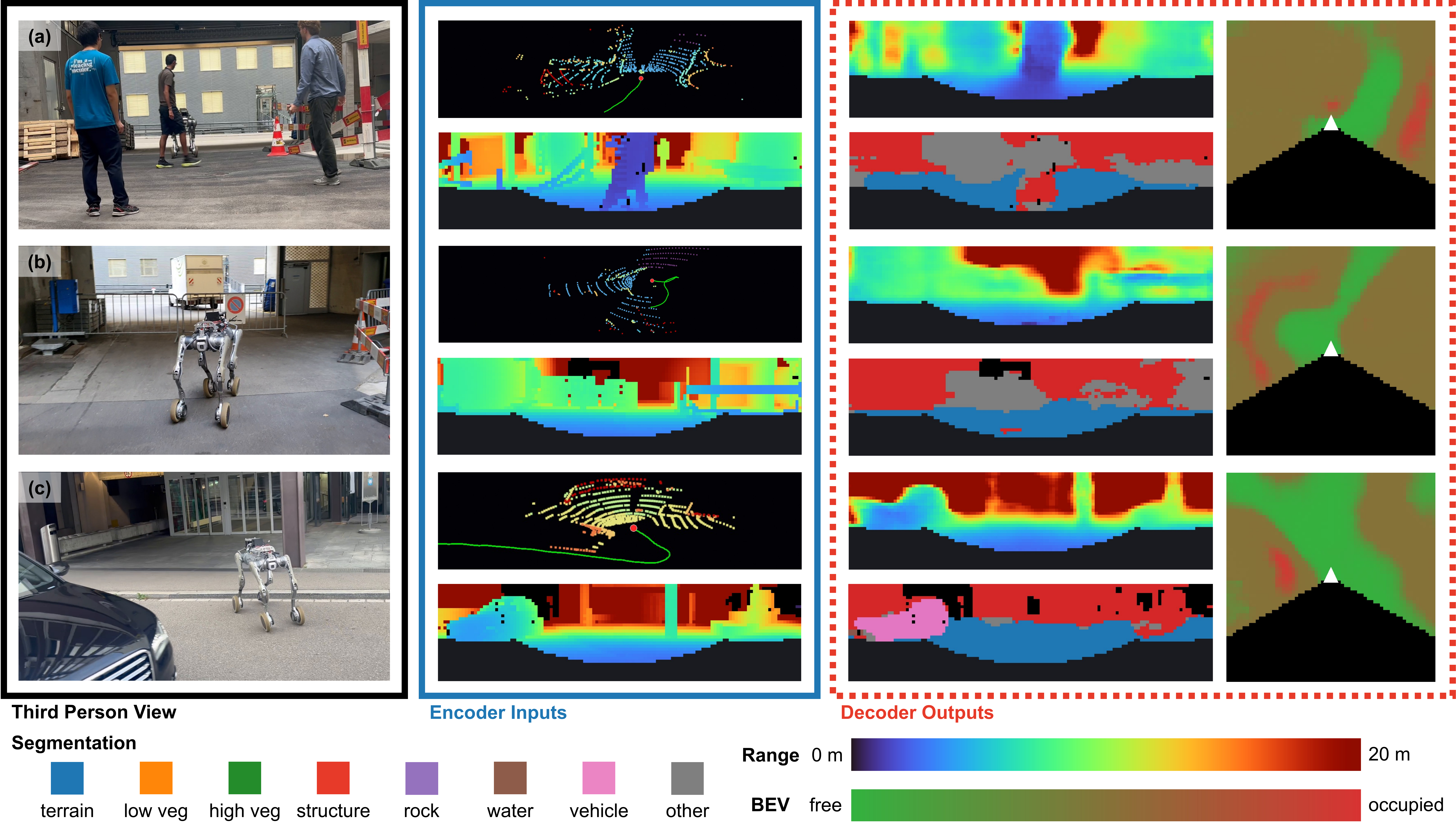}
    \caption{Perception examples from physical deployment with (a) pedestrians, (b) a fence, and (c) vehicles. Each row shows the scene photograph, the input point cloud and its projected range image (blue box, top and bottom), and the decoder outputs (red dashed box): range reconstruction (left, top), semantic segmentation (left, bottom), and BEV prediction (right).}
    \label{fig:perception_examples}
    \vspace{-1em}
\end{figure*}

Transferring a navigation policy trained exclusively in simulation requires robustness to real-world sensor noise and occlusion, unstructured geometry, and dynamic obstacles. We evaluate whether our perception and navigation pipeline can handle these conditions without real-world fine-tuning.

Our hardware platform is a Unitree B2-W equipped with a RoboSense Helios 32-beam LiDAR and an IMU. BIEVR-LIO~\cite{pfreundschuh2026bievr} estimates the robot pose from the LiDAR and inertial measurements. The full stack runs onboard an NVIDIA Jetson AGX Orin. Because a rear-mounted payload box occludes part of the LiDAR scan, we mask the rear $120^\circ$ sector before projecting each scan into a range image, as indicated by the black regions in Fig.~\ref{fig:perception_examples}. The encoder then maps this range image to a latent fed into the RL policy. Fig.~\ref{fig:perception_examples} visualizes the learned representation through the reconstruction, semantic segmentation, and BEV prediction decoder outputs. These outputs show accurate geometry reconstruction, segmentation of obstacles such as fences and cars, and BEV occupancy prediction.

We evaluate the system across diverse real-world environments that challenge both sensing and navigation. In low-light and dark scenes, LiDAR provides geometric observations independently of ambient illumination. Dynamic crowds test whether the navigation policy can respond online to pedestrians whose positions and trajectories continually change. In construction sites with dense clutter, narrow passages, and irregular geometry, panoramic LiDAR reveals obstacles and openings outside a front-facing depth camera's field of view, while the recurrent policy retains temporal context to avoid revisiting dead ends. Representative deployments are shown in Figs.~\ref{fig:overview} and~\ref{fig:perception_examples}. Successful rollouts across these settings show that the LiDAR-based system can navigate under limited illumination while its mapless policy balances reactive pedestrian avoidance with longer-horizon progress, continuing through temporarily obstructed passages once they clear while avoiding repeated attempts at impassable routes. Because the same perception--policy interface is used in simulation and on the robot, these rollouts also demonstrate zero-shot sim-to-real transfer.

\subsection{Limitations}

RangeNav remains limited by sensor visibility and sensor-specific scan geometry. Sparse LiDAR sampling, range-image projection and downsampling, and sensor occlusions can discard fine structure or leave thin and low-lying obstacles unobserved. The rear payload mask further creates a substantial blind sector on the deployed platform. Moreover, the fixed-size range image is tied to the channel count, field of view, and angular resolution of the target LiDAR. Deployment with a different scan pattern may therefore require explicit resampling before encoding or pretraining the encoder from scratch with sensor-specific input dimensions. Dynamic obstacles are also currently handled reactively, without explicit agent tracking or motion prediction, making sudden interactions and dense crowds potential failure cases. Future work will address cross-sensor adaptation and explicit motion prediction.

\section{Conclusion}

We present a compact, task-aligned LiDAR range-image encoder for RL navigation. Using synthetic range images with segmentation and BEV labels, we pretrain an architecture tailored to the spatial structure of LiDAR observations. Adding semantic segmentation and BEV prediction objectives substantially improves downstream navigation performance over range reconstruction alone. The resulting encoder supports efficient policy learning, and its task-aligned features yield better navigation performance than those distilled from substantially larger general-purpose point-cloud encoders. Compared with a front-facing depth policy, our LiDAR policy achieves greater path efficiency, reaches goals faster, and performs better in cluttered and dynamic-obstacle settings. Finally, deployment on a physical robot demonstrates zero-shot transfer in dark, crowded, and construction environments. Taken together, these findings establish task-aligned range-image pretraining as a practical foundation for LiDAR navigation.

\section*{\small ACKNOWLEDGEMENTS}
{\small
\anon{This research was supported by the Swiss National Science Foundation (SNSF) under Project No.~227617 and Grant 200021E\_229503, the Luxembourg National Research Fund (Ref.~18990533), and armasuisse Science and Technology.\par}
The author(s) used generative AI to assist with drafting and editing
the manuscript. All content was reviewed by the authors, who take
full responsibility for it.
\par}

\bibliographystyle{IEEEtran}
\bibliography{ref}

\end{document}